\documentclass[review]{elsarticle}

\usepackage{lineno,hyperref}
\modulolinenumbers[5]

\journal{Neural Networks}

\usepackage{graphicx}
\usepackage{subfigure}
\usepackage{booktabs} 
\usepackage{multirow}

\usepackage{amsmath}
\usepackage{amssymb}
\usepackage{amsthm}
\graphicspath{ {./figure/} }
\numberwithin{equation}{section}

\usepackage{mathtools}
\newcommand\coolover[2]{\mathrlap{\smash{\overbrace{\phantom{%
    \begin{matrix} #2 \end{matrix}}}^{\mbox{$#1$}}}}#2}

\newcommand\coolrightbrace[2]{%
\left.\vphantom{\begin{matrix} #1 \end{matrix}}\right\}#2}

\newcommand\sumk{\sum_{k=1}^N}
\newcommand\tR{\widetilde{R}}
\newcommand\bR{\bar{R}}
\newcommand\bbP{\mathbb{P}}
\newcommand\gp{\mathrm{gp}}

\newcommand\proto{\mathrm{proto}}
\renewcommand\L{\mathrm{L}}

\DeclareMathOperator{\diag}{diag}
\DeclareMathOperator{\ran}{ran}

\begin{document}

\begin{frontmatter}

\title{Two-level Group Convolution}

\author[mymainaddress]{Youngkyu Lee}
\ead{lyk92@kaist.ac.kr}

\author[mysecondaryaddress]{Jongho Park}
\ead{jongho.park@kaist.ac.kr}

\author[mymainaddress]{Chang-Ock Lee\corref{mycorrespondingauthor}}
\cortext[mycorrespondingauthor]{Corresponding author}
\ead{colee@kaist.edu}

\address[mymainaddress]{Department of Mathematical Sciences, KAIST, Daejeon 34141, Korea}
\address[mysecondaryaddress]{Natural Science Research Institute, KAIST, Daejeon 34141, Korea}

\begin{abstract}
    Group convolution has been widely used in order to reduce the computation time of convolution, which takes most of the training time of convolutional neural networks.
    However, it is well known that a large number of groups significantly reduce the performance of group convolution.
    In this paper, we propose a new convolution methodology called ``two-level'' group convolution that is robust with respect to the increase of the number of groups and suitable for multi-GPU parallel computation.
    We first observe that the group convolution can be interpreted as a one-level block Jacobi approximation of the standard convolution, which is a popular notion in the field of numerical analysis.
    In numerical analysis, there have been numerous studies on the two-level method that introduces an intergroup structure that resolves the performance degradation issue without disturbing parallel computation.
    Motivated by these, we introduce a coarse-level structure which promotes intergroup communication without being a bottleneck in the group convolution.
    We show that all the additional work induced by the coarse-level structure can be efficiently processed in a distributed memory system.
    Numerical results that verify the robustness of the proposed method with respect to the number of groups are presented.
    Moreover, we compare the proposed method to various approaches for group convolution in order to highlight the superiority of the proposed method in terms of execution time, memory efficiency, and performance.
\end{abstract}


\end{frontmatter}

\section{Introduction}
\label{Sec:Int}
Modern convolutional neural networks~(CNNs) have become deeper and larger to improve their performance~\cite{he2016deep,he2016identity,krizhevsky2012imagenet,long2015fully,szegedy2015going}.
Almost all successful CNNs consist of hundreds of layers and thousands of channels~\cite{he2016identity,szegedy2016rethinking,xie2017aggregated}, which require significant computational cost to handle them.
Increased computational cost has led to increased training time, and numerous research has been conducted to shorten the training time by modifying the convolutions in CNNs; see, e.g.,~\cite{huang2018condensenet,ioannou2017deep,wang2019fully,zhang2018shufflenet}.

Group convolution is a simple and basic modification of convolution.
It was first proposed in AlexNet~\cite{krizhevsky2012imagenet} for distributed computing of convolutions in CNN over multiple GPUs.
It was shown in~\cite{xie2017aggregated,huang2018condensenet,ioannou2017deep,chollet2017xception} that the group convolution is very effective on reducing both the number of parameters and  the training time of CNN.
However, it has a drawback that the accuracy of CNN equipped with group convolutions severely decreases when the number of groups is large.
This is because groups in the group convolution do not communicate with each other, which greatly affects the representation capacity of the CNN.

Recently, several group convolution techniques that add intergroup communication have been considered to prevent the decrease of performance.
A channel shuffling technique~\cite{zhang2018shufflenet} adds a permutation step among the output channels of the group convolution in order to make data exchange among groups.
A fully learnable group structure~\cite{wang2019fully} optimizes the group arrangement of the input and output channels of the group convolution.
Depthwise separable convolutions proposed in~\cite{chollet2017xception} improve computational performance and reduce the number of parameters by distributing the channelwise and spatialwise computations of the standard convolution into two convolutions.
Although these works successfully resolved the performance degradation issue explained above, the parallel structure of the group convolution is disturbed by the additional structures, hence, parallel implementation of these methods in a distributed memory system becomes difficult.
Therefore, it is important to develop a novel group convolution that resolves the performance degradation issue while preserving the parallel structure.
This paper is devoted to such a novel group convolution called \textit{two-level} group convolution.

In view of numerical analysis, the group convolution can be regarded as a block Jacobi approximation of the standard convolution.
Block Jacobi method in numerical analysis is one of the basic parallel algorithms for scientific computing~(see, e.g.,~\cite{lee2019fast,saad2003iterative}) and shares the same drawback as the group convolution.
That is, as the number of groups increases, the quality of block Jacobi approximation decreases.
However, there is a popular remedy called two-level correction for the disadvantage of the block Jacobi method~\cite{dolean2015introduction,toselli2005domain}.
In two-level methods, a coarse-level grid is added for the purpose of intergroup communication.
Usually, it covers the whole domain of the target problem but with a less number of degrees of freedom so that it provides data exchange among the whole domain with marginal computational cost.
Due to its small computational cost, it does not harm the parallel structure of the algorithm.
Such a two-level idea has been successfully applied to various problems in numerical analysis; see~\cite{dolean2015introduction,toselli2005domain}.

Motivated by such two-level methods, we present an additional structure to the group convolution that grants communication among groups with minimal computational cost and without breaking the parallel structure.
We call the group convolution plus such an additional structure two-level group convolution.
In the two-level group convolution, a single channel representating each group is generated through groupwise computation, and then communicates so that all groups share the information of all representatives.
After that, the information of such representative channels are added to the output of the group convolution in an appropriate manner.
Compared to existing approaches, the proposed two-level group convolution maintains the parallel structure in the sense that no communication of parameters is required; only a small number of data communication is required.
Numerical results ensure that the proposed two-level group convolution provides better performance compared to existing works in terms of suitability for parallel computation and robustness with respect to the increase of the number of groups.

The rest of this paper is organized as follows.
In Section~\ref{Sec:Jacobi}, we introduce an abstract block Jacobi framework and express the group convolution as a block Jacobi approximation of the standard convolution.
We present the proposed two-level group convolution motivated by a two-level extension of the block Jacobi method in Section~\ref{Sec:Twolevel}.
Improved classification accuracy of the proposed two-level group convolution applied to WideResNet~\cite{zagoruyko2016wide} and MobileNetV2~\cite{sandler2018mobilenetv2} with various datasets are presented in Section~\ref{Sec:App}.
We conclude this paper with remarks in Section~\ref{Sec:Conc}.

\section{Group convolution as block Jacobi approximation}
\label{Sec:Jacobi}
In this section, we introduce an algebraic framework of block Jacobi approximation for linear operators.
Then we represent group convolution in the framework of block Jacobi approximation.
In this perspective, we will adopt concepts from numerical analysis in order to improve group convolution in the next section.

\subsection{General block Jacobi approximation}
Let $V$ and $W$ be real vector spaces and $N$ be a positive integer.
We assume that there exist a vector space $V_k$ and a linear operator $R_k \colon V \rightarrow V_k$ such that
\begin{equation}
\label{V_decomp}
V = \sumk R_k^T V_k.
\end{equation}
Similarly, we assume that a vector space $W_k$ and a linear operator $\tR_k \colon W \rightarrow W_k$ satisfy
\begin{equation}
\label{W_decomp}
W = \sumk \tR_k^T W_k.
\end{equation}

We construct a block Jacobi approximation of a linear operator $A \colon V \rightarrow W$ under the space decomposition settings~\eqref{V_decomp} and~\eqref{W_decomp}.
A local operator $A_k \colon V_k \rightarrow W_k$ is defined by
\begin{equation}
\label{local_op}
A_k = \tR_k A R_k^T, \quad 1 \leq k \leq N.
\end{equation}
Here, $R_k^T$ plays a role of prolongation that embeds an element of the local space $V_k$ into the global space $V$, while $\tR_k$ restricts an element of the global space $W$ to the local space $W_k$.
The \textit{block Jacobi approximation} $M$ of $A$ with local operators~\eqref{local_op} is given by
\begin{equation}
\label{Jacobi}
M = \sumk \tR_k^T A_k R_k.
\end{equation}
Operators of the form~\eqref{Jacobi} are typical in the field of numerical analysis; see, e.g.,~\cite{lee2019fast,saad2003iterative}.
As a descriptive example, we consider the case when the decompositions~\eqref{V_decomp} and~\eqref{W_decomp} are of direct sums, i.e.,
\begin{equation}
\label{direct_sum}
V = \bigoplus_{k=1}^N V_k \quad\textrm{ and }\quad W = \bigoplus_{k=1}^N W_k.
\end{equation}
In view of domain decomposition methods, space decompositions in~\eqref{direct_sum} are called \textit{one-level} since subspaces $V_k$ and $W_k$ correspond to subdomains when $V$ and $W$ are appropriate function spaces defined on entire domains; see~\cite{toselli2005domain}.

For the block matrix representation
\begin{equation}
\label{A_matrix}
A = \left[ A_{ij} \right]_{1 \le i,j \le N}
= \begin{bmatrix}
A_{11} & A_{12} & \dots & A_{1N} \\
A_{22} & A_{22} & \dots & A_{2N} \\
\vdots & \vdots & \ddots & \vdots \\
A_{N1} & A_{N2} & \dots & A_{NN}
\end{bmatrix},
\end{equation}
with respect to the decompositions~\eqref{direct_sum}, the corresponding block matrix representation of the block Jacobi approximation $M$ is written as
\begin{equation}
\label{M_matrix}
M = \diag \left( \left[ A_{ii} \right]_{i=1}^N \right)
= \begin{bmatrix}
A_{11} & 0 & \dots & 0 \\
0 & A_{22} & \dots & 0 \\
\vdots & \vdots & \ddots & \vdots \\
0 & 0 & \dots & A_{NN}
\end{bmatrix},
\end{equation}
where $A_{ij} \colon V_j \rightarrow W_i$ is defined by $A_{ij}=\tR_i A R_j^T$.
That is, $M$ is the block-diagonal part of $A$.
Comparing~\eqref{A_matrix} and~\eqref{M_matrix}, it is clear that as $N$ increases, the quality of approximation of $M$ with one-level space decompositions decreases significantly; see~\cite{toselli2005domain}.

\subsection{Group convolution}
Now, we show that the group convolution proposed in~\cite{krizhevsky2012imagenet} fits the framework of the block Jacobi approximation.
For convenience, it is assumed that all images considered in this section are of the same size, with height and width of $h$ and $w$, respectively.
Let $\bbP$ be the vector space consisting of all grayscale images, i.e., $\bbP = \mathbb{R}^{h \times w}$.
For a positive integer $n$, we may write the space of all $n$-channel images as $\bbP^n$.
In the following, let $C_d \colon \bbP \rightarrow \bbP$ denote a generic $d \times d$ convolution on $\bbP$.

The full $d \times d$ convolution $A \colon \bbP^n \rightarrow \bbP^m$ from $n$-channel images to $m$-channel images can be written in a block form
\begin{equation}
\label{full_block}
A = \left[ C_d \right]_{m \times n}
=
\vphantom{
    \begin{matrix}
    \overbrace{dummy}^{dummy}\\ \\ \\
    \end{matrix}}%
\begin{bmatrix}
\coolover{\scriptstyle{n} \textrm{  times}}{C_d & C_d & \dots & C_d} \\
C_d & C_d & \dots & C_d \\
\vdots & \vdots & \ddots & \vdots \\
C_d & C_d & \dots & C_d
\end{bmatrix}%
\begin{matrix}
    \coolrightbrace{x \\ x \\ y\\ y}{\scriptstyle{m} \textrm{ times}}
\end{matrix}.
\end{equation}
Now, let $N$ be a common divisor of $m$ and $n$.
We consider a $d \times d$ group convolution $A_{\gp} \colon \bbP^n \rightarrow \bbP^m$ with $N$ groups so that $n$ input and $m$ output channels are partitioned into $N$ groups of $n/N$ and $m/N$ channels, respectively.
Then $N$ convolution operations from $n/N$-channel images to $m/N$-channel images are done separately.
If we express the group convolution $A_{\gp}$ in a compact form, we have
\begin{equation}
\label{group_block}
A_{\gp} = \diag \left( \left[ \left[ C_d \right]_{m/N \times n/N} \right]_{k=1}^N \right).
\end{equation}
Note that, by assembling block matrices appropriately, the standard convolution in~\eqref{full_block} is rewritten as
\begin{equation}
\label{full_block_re}
A = \left[ \left[C_d \right]_{m/N \times n/N} \right]_{N \times N}.
\end{equation}
Observing that~\eqref{group_block} and~\eqref{full_block_re} have the same forms as~\eqref{M_matrix} and~\eqref{A_matrix}, respectively, we can say that the group convolution $A_{\gp}$ is indeed a one-level block Jacobi approximation of the standard convolution $A$.

Equation~\eqref{group_block} can be written in a more abstract fashion.
We set $V = \bbP^n$, $W = \bbP^m$, $V_k = \bbP^{n/N}$, and $W_k = \bbP^{m/N}$.
Let $R_k \colon V \rightarrow V_k$ and $\tR_k \colon W \rightarrow W_k$ be the restriction operators onto the $k$th groups.
Then it is straightforward to observe that $A_{\gp}$ agrees with the corresponding block Jacobi approximation $M$ in~\eqref{Jacobi}, i.e.,
\begin{equation}
\label{gp_1L}
A_{\gp} = \sumk \tR_k^T A_k R_k,
\end{equation}
where the local operator $A_k \colon V_k \rightarrow W_k$ defined in~\eqref{local_op} is given by
\begin{equation}
\label{local_op_gp}
A_k = [C_d]_{m/N \times n/N}.
\end{equation}
We note that entries of $R_k$ and $\tR_k$ consist of 0's and 1's only, while $A_k$ is composed of parameters to be determined by training.

\section{Two-level group convolution}
\label{Sec:Twolevel}
As we observed in Section~\ref{Sec:Jacobi}, the group convolution $A_{\gp}$ is an instance of one-level block Jacobi approximations of the standard convolution $A$.
Since the quality of approximation of one-level block Jacobi approximation is deteriorated when $N$ becomes large, it can be expected that the performance of neural networks using group convolution heavily depends on the number of groups $N$; see~\cite{wang2019fully} for relevant numerical results.
There have been several works~\cite{wang2019fully,zhang2018shufflenet} to address these shortcomings of group convolution, but all of these harm the parallel structure, another desirable property of group convolution.
\subsection{Algebraic framework}
\label{Subsec:Algebraic}
On the other hand, in the numerical analysis literature, there is a standard treatment called a \textit{two-level} method to compensate for the above-mentioned problem of the one-level method without disturbing the parallel structure; see, e.g.,~\cite{dolean2015introduction,toselli2005domain}.
In two-level methods, an additional subspace $V_0$ called \textit{coarse space} is added in the decomposition~\eqref{V_decomp} so that we get
\begin{equation}
\label{V_decomp_2L}
V = R_0^T V_0 + \sumk R_k^T V_k,
\end{equation}
where $R_0 \colon V \rightarrow V_0$ is a suitable linear operator.
Usually, the basis of $V_0$ contains $O(N)$ elements, where each element plays a role of a representative of the subspace $V_k$.
Therefore, the coarse space $V_0$ promotes communication between local subspaces $V_k$'s.
While computations regarding the coarse space $V_0$ cannot be parallelized due to its global nature, they do not disturb the parallel structure of the method severely since the dimension of $V_0$ is $O(N)$, which is almost negligible compared to the dimension of the entire space $V$.
The same idea applies to the space $W$, so we have
\begin{equation*}
W = \tR_0^T W_0 + \sumk \tR_k^T W_k,
\end{equation*}
where $\tR_0 \colon W \rightarrow W_0$ is a linear operator.

\begin{figure}
\begin{center}
\includegraphics[width=0.5\linewidth, trim=6cm 1cm 6cm 1cm,clip]{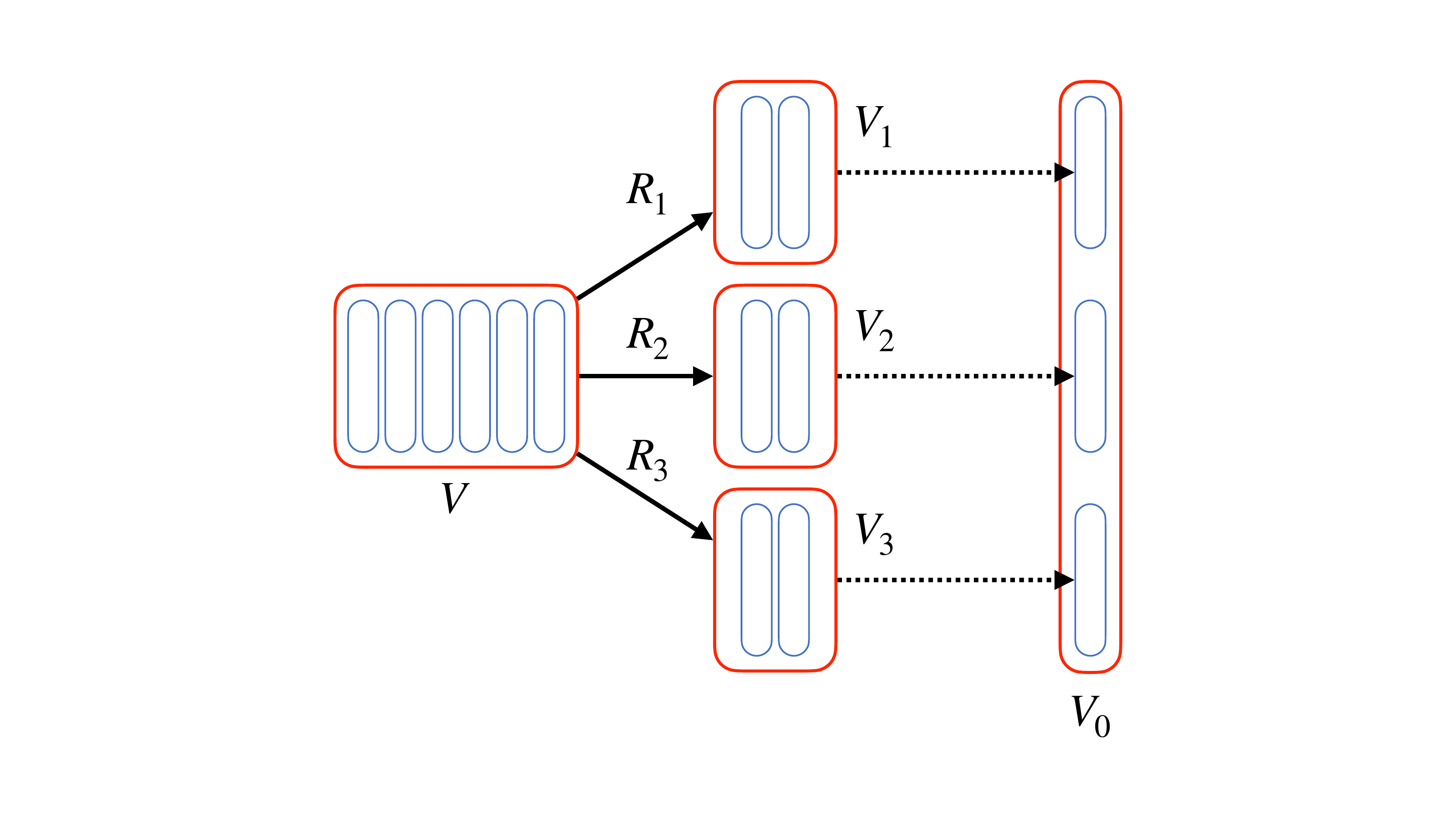}
\end{center}
\caption{A graphical description of the coarse space $V_0$ for $N=3$. The entire space $V$ is decomposed into subspace $V_i$ for $i=1,2,3$ satisfying $V=\bigoplus_{i=1}^{3}V_i$. The elements of the coarse space $V_0$ are representatives of each subspace $V_i$.}
\label{coarsespace}
\end{figure}

Motivated by the two-level method, we propose a novel group convolution called \textit{two-level group convolution}, which preserves the parallel structure without significantly affecting the performance by increasing the number of groups.
Beginning with~\eqref{V_decomp_2L}, let $V_0$ be the space of $N$-channel images in which each channel plays a role of a representative of the group $V_k$.
Likewise, we define $W_0$ as the space of $N$-channel images; see Figure~\ref{coarsespace} for a graphical description of the coarse spaces.
For suitable operators $A_0 \colon V_0 \rightarrow W_0$, $R_0 \colon V \rightarrow V_0$, and $\tR_0 \colon W \rightarrow W_0$ that will be defined later, a prototype $A_{2\L, \proto}$ for the two-level group convolution is constructed by adding a coarse correction term $\tR_0^T A_0 R_0$ to~\eqref{gp_1L} as follows:
\begin{equation}
\label{gp_2L}
A_{2\L, \proto} = \tR_0^T A_0 R_0 + \sumk \tR_k^T A_k R_k.
\end{equation}
While the restriction operators $R_0$ and $\tR_0$ are naturally defined according to the geometry of subspaces in the case of classical domain decomposition methods~\cite[Sect.~3.3]{toselli2005domain}, we have to consider carefully how to define these operators in the case of group convolution since there are no underlying geometries for the coarse spaces $V_0$ and $W_0$.
In the following, we specify the coarse space $V_0$ and the operators $A_0$, $R_0$, and $\tR_0$ in the coarse correction term of~\eqref{gp_2L}.

As mentioned above, the operator $R_0$ must be designed so that, for any $x \in V$, the $k$th channel of $R_0 x \in V_0$ contains information of the whole $V_k$.
In this sense, we define $R_0$ so that the $k$th channel $(R_0 x)_k$ of $R_0 x$ is constructed by the convolution applied to $R_k x$ consisting of $(n/N)$ channels, i.e.,
\begin{equation}
\label{R0}
(R_0 x)_k = [C_{d_0}]_{1 \times n/N} R_k x, \quad x \in V,
\end{equation}
where $d_{0}$ is a positive integer.
As we will see in Section~\ref{Subsec:Implementation}, a suitable choice for the kernel size $d_{0}$ is $d$, the size of the kernel $C_d$ in~\eqref{full_block}.
It is observed from~\eqref{R0} that each channel $(R_0 x)_k$ can be computed independently.

The main purpose of the coarse operator $A_0$ is data communication between groups.
It is desired that, for any $x_0 \in V_0$, each channel of $A_0 x_0 \in W_0$ is constructed by using all channels of $x_0$.
At the same time, the computational cost of $A_0$ should be as cheap as possible to avoid becoming a bottleneck for parallel computation; it is also an important issue to reduce the size of coarse structures in classical domain decomposition methods~\cite{dohrmann2017design,klawonn2015feti}.
A reasonable choice for $A_0$ that meets both of the above requirements is the full $1 \times 1$ convolution from $V_0$ to $W_0$:
\begin{equation*}
\label{A_0}
A_0 = [C_1]_{N \times N}.
\end{equation*}

The operator $\tR_0^T$ distributes each channel $(y_0)_k$ of $y_0 \in W_0$ to $W_k$ in an appropriate manner.
We simply set $\tR_0^T$ as the combination of $1 \times 1$ convolutions from each channel $(y_0)_k$ of $y_0$ to $m/N$ channels in $W_k$.
That is, we set
\begin{equation*}
\label{tR_0}
\tR_0^T y_0 = \sumk \tR_k^T [C_1]_{m/N \times 1} (y_0)_k, \quad y_0 \in W_0.
\end{equation*}

Now, $A_0$, $R_0$, and $\tR_0^T$ in $A_{2\L, \proto}$ are all specified.
In addition, by simplifying the coarse structure, we construct the final version $A_{2\L}$ of two-level group convolution.
Note that $1 \times 1$ convolutions are linear combinations of channels.
Therefore, for any $x_0 \in V_0$, every channel of $A_0 x_0 \in  W_0$ is a linear combination of all channels of $x_0$.
Similarly, all channels in the $k$th group of $\tR_0^T A_0 x_0 \in W$ are scalar multiples of the $k$th channel of $A_0 x_0$.
Therefore, we can see that $\tR_0^T A_0 x_0$ is composed of linear combinations of all channels of $x_0$.
In this perspective, the proposed two-level group convolution $A_{2\L}$ is made by replacing $\tR_0^T A_0$ in the coarse structure of $A_{2\L,\proto}$ by the full $1 \times 1$ convolution from $V_0$ to $W$, say $\bR_0^T \colon V_0 \rightarrow W$ such that
\begin{equation}
\label{bR0}
\bR_0^T = \sumk \tR_k^T [C_1]_{m/N \times N} = \sumk \tR_k^T S_k,
\end{equation}
where $S_k \colon V_0 \rightarrow W_k$ is a $1 \times 1$ standard convolution from $V_0$ to $W_k$.
There are two advantages of replacing $\tR_0^T A_0$ by $\bR_0^T$.
The first one is obvious;  the coarse structure of $A_{2\L}$ is simpler to implement than $A_{2\L, \proto}$.
Moreover, $\ran \bR_0^T$ is larger than $\ran \tR_0^T A_0$ in general.
As a result, the performance improvement can be expected as the number of parameters increases.

Note that each component $\tilde{R}_k^T [C_1]_{m/N \times N}$ in~\eqref{bR0} can be processed in parallel if every processor possesses the whole $N$ channels of an element of $V_0$.
Since the $k$th processor owns the $k$th channel of $V_0$ after parallel computation of $R_0$, data communication among processors is required before applying $\bR_0^T$; details will be given in Section~\ref{Subsec:Implementation}.

Finally, we summarize the proposed two-level group convolution as follows:
\begin{equation}
\label{proposed}
A_{2\L} = \bR_0^T R_0 + \sumk \tR_k^T A_k R_k = \bR_0^T R_0 + A_{\gp}.
\end{equation}

\subsection{Implementation issues}
\label{Subsec:Implementation}

We discuss several issues on efficient parallel implementation of the proposed two-level group convolution.
First, we evaluate how many parameters are dealt with by a single processor in parallel implementation of the two-level group convolution using $N$ processors.
In~\eqref{proposed}, the two-level group convolution $A_{2\L}$ consists of two parts: the group convolution $A_{\gp} = \sumk \tR_k^T A_k R_k$ and the coarse structure $\bR_0^T R_0$.
Since $R_k$ and $\tR_k^T$ consist of 0's and 1's and do not contain any parameters, it suffices to consider the number of parameters in $A_k$, $R_0$, and $\bR_0^T$ only.
Note that both $R_k$ and $\tR_k^T$ do not need to be explicitly assembled in implementation.
The group convolution $A_{\gp}$ is composed of $N$ parallel blocks of $A_k = [C_d]_{m/N \times n/N}$ as shown in~\eqref{group_block} and~\eqref{local_op_gp}.
Similarly, $R_0$ and $\bR_0^T$ are composed of $N$ blocks of $[C_{d_{0}}]_{1 \times n/N}$ and $[C_1]_{m/N \times N}$ as shown in~\eqref{R0} and~\eqref{bR0}, respectively.
If each processor holds the parameters of $[C_d]_{m/N \times n/N}$ and $[C_{d_0}]_{1 \times n/N}$, then no communication of parameters among processors are required.
That is, the number of parameters per processor of the proposed two-level group convolution is given by
\begin{equation*}
\label{parameters}
d^2 \cdot \frac{m}{N} \cdot \frac{n}{N} + d_{0}^2 \cdot \frac{n}{N} = \frac{d^2 mn}{N^2} + \frac{d_{0}^2 n}{N}.
\end{equation*}
On the other hand, the number of parameters of the coarse structure $\bR_0^T$ is $mN$.
During the two-level computation, data communication between processors is required even though the distributed parameters do not need to be communicated.
Since parallel computation of $R_0$ is done in a way that the $k$th processor deals with the channels of $V_k$ to generate the $(R_{0}x)_{k}$, each processor has only a single channel of $R_{0}x$ just after an application of $R_0$.
After that, a collective interprocess operation gathers $(R_{0}x)_{k}$ from each processor and then apply $\bR_0^T$ to $R_{0}x \in V_{0}$.
In the application of coarse computation $\bR_0^T$, we may assign a single dedicated processor to handle whole $mN$ parameters of $\bR_0^T$ or use all processors where $\tR_k^T[C_{1}]_{m/N \times N}$ is applied separately.
In the latter, each processor deals with additional $m$ parameters so that we can avoid the coarse structure being a bottleneck when $N$ is large.
Also it makes the number of parameters per processor of the two-level group convolution slightly more than that of group convolution.
We chose the latter for our implementation.

\begin{figure}
\begin{center}
\includegraphics[width=0.8\linewidth, trim=0cm 1cm 0cm 1cm,clip]{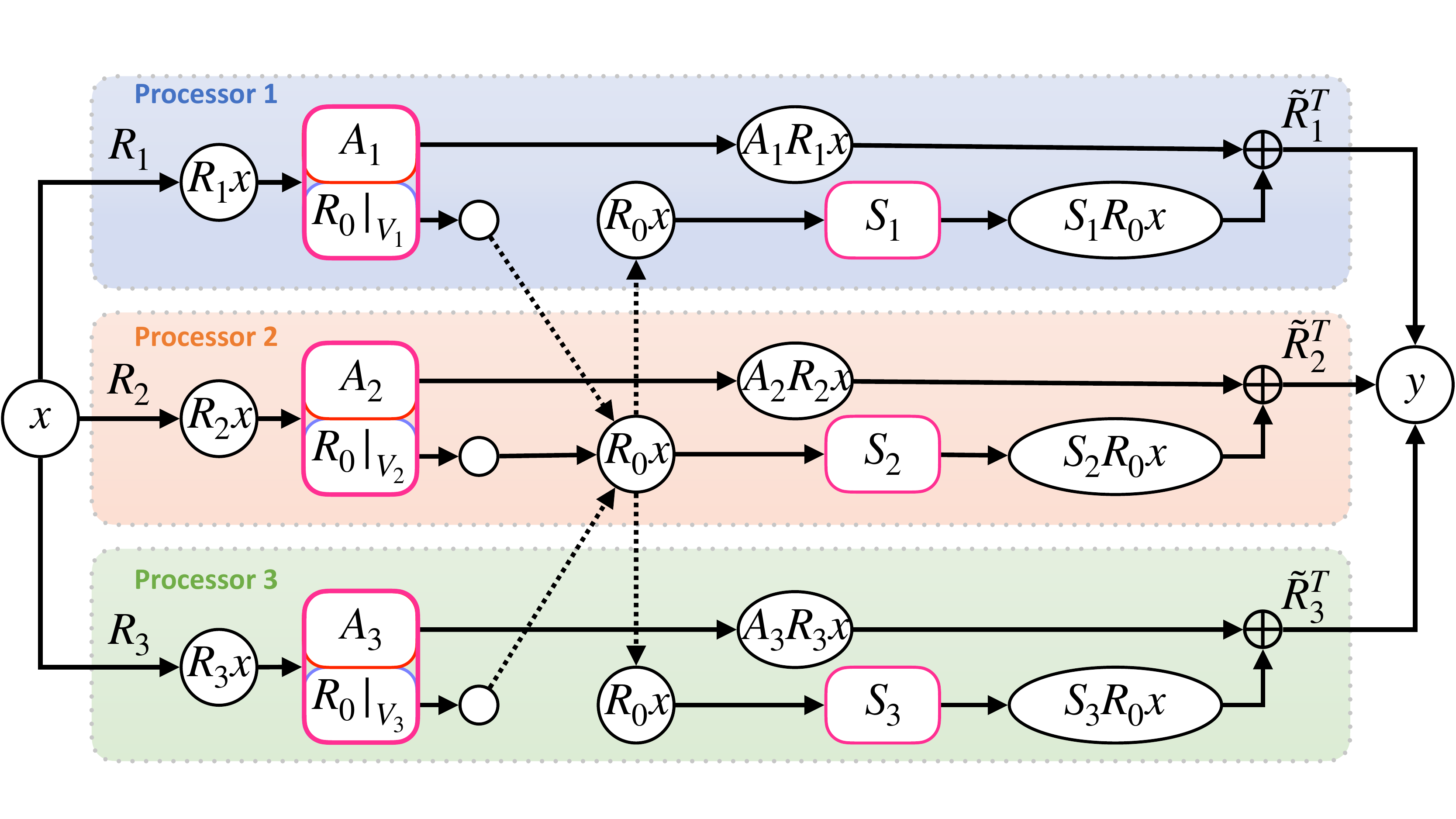}
\end{center}
\caption{Overall structure of the proposed two-level group convolution in the implementation perspective when $N=3$.
Note that $S_{k}=[C_{1}]_{m/N \times N}$.
Each processor has no parameter communication and only the representative channels of the groups communicate.
The dotted lines denotes such communications.}
\label{twolevel}
\end{figure}

Next, we discuss why the choice $d_{0}= d$ of the kernel size of $R_0$ in~\eqref{R0} is suitable.
In the implementation of the two-level group convolution, applications of $A_k$ and $R_0$ can be done simultaneously since $R_k$ has no computational cost.
If $d_{0}=d$, $A_k$ and the restriction $R_0|_{V_k}$ of $R_0$ onto $V_k$ can be implemented in combination as
\begin{equation*}
\label{combine}
\begin{bmatrix}A_k \\ R_0 |_{V_k}\end{bmatrix} = [C_d]_{(m/N+1) \times n/N}.
\end{equation*}
That is, $A_{\gp}$ and $R_0$ can be implemented simultaneously by a convolution from $n/N$ channels to $(m/N + 1)$ channels in each processor if $d_{0} = d$.
Therefore, we choose $d_{0}= d$ for efficient parallel implementation.
We also note that, in general, a combined implementation of two convolutions is more efficient than separate implementations since convolutions are implemented by utilizing fast Fourier transform in libraries for scientific computing; see, e.g.,~\cite{chetlur2014cudnn}.
The overall procedure of the proposed two-level group convolution is depicted in Figure~\ref{twolevel}.

\section{Applications}
\label{Sec:App}
In this section, we present numerical results of the proposed two-level group convolution embedded into existing CNNs: WideResNet~\cite{zagoruyko2016wide} and MobileNetV2~\cite{sandler2018mobilenetv2}.
In the experiements with WideResNet, we discuss the classification performance and parallel efficiency of the two-level group convolution based on the results applied to $3 \times 3$ convolutions.
On the other hand, in the experiments with MobileNetV2, based on the results applied to $1 \times 1$ convolutions, we show that the number of network parameters is effectively reduced while minimizing the performance degradation compared to the group convolution and its variants~\cite{wang2019fully,zhang2018shufflenet}.

We implemented our method on datasets CIFAR-10, CIFAR-100~\cite{krizhevsky2009cifar}, and ImageNet ILSVRC 2012~\cite{deng2009imagenet}.
Details on the datasets we used are as follows.
The CIFAR-$m$~($m=10, 100$) dataset consists of $32 \times 32$ colored natural images including 50,000 training and 10,000 test samples with $m$ classes.
ImageNet is a dataset which contains 1,000 classes of $224 \times 224$ colored natural images.
It includes 1,280,000 training and 50,000 test samples.

For CIFAR datasets, a data augmentation technique in~\cite{lee2015deeply} was adopted for training; four pixels are padded on each side of images and $32 \times 32$ random crops are sampled from the padded images and their horizontal flips.
All neural networks for CIFAR datasets in this section were trained using the stochastic gradient descent with the batch size 128, weight decay $0.0005$, momentum $0.9$, total epoch 200, and weights initialized as in~\cite{he2015delving}.
The initial learning rate was set to $0.1$ and was reduced to its one tenth in the 60th, 120th, and 160th epochs.

For ImageNet dataset, the input image is $224 \times 224$ randomly cropped from a resized image using the scale and aspect ratio augmentation of~\cite{szegedy2015going} implemented by~\cite{gross2016training}.
Hyperparameter settings are the same as the CIFAR case except the followings: the weight decay $0.0001$, total epoch $90$, and the learning rate reduced by a factor of 10 in the 30th and 60th epochs.

All networks were implemented in Python with PyTorch~\cite{paszke2019pytorch} and all computations were performed on a cluster equipped with Intel Xeon Gold 5515 (2.4GHz, 20C) CPUs, NVIDIA Titan RTX GPUs, and the operating system Ubuntu 18.04 64bit.

\subsection{WideResNet}
\label{Sec:Wide}
\begin{figure}
\begin{center}
\includegraphics[width=0.8\linewidth, trim=0cm 0cm 0cm 0cm,clip]{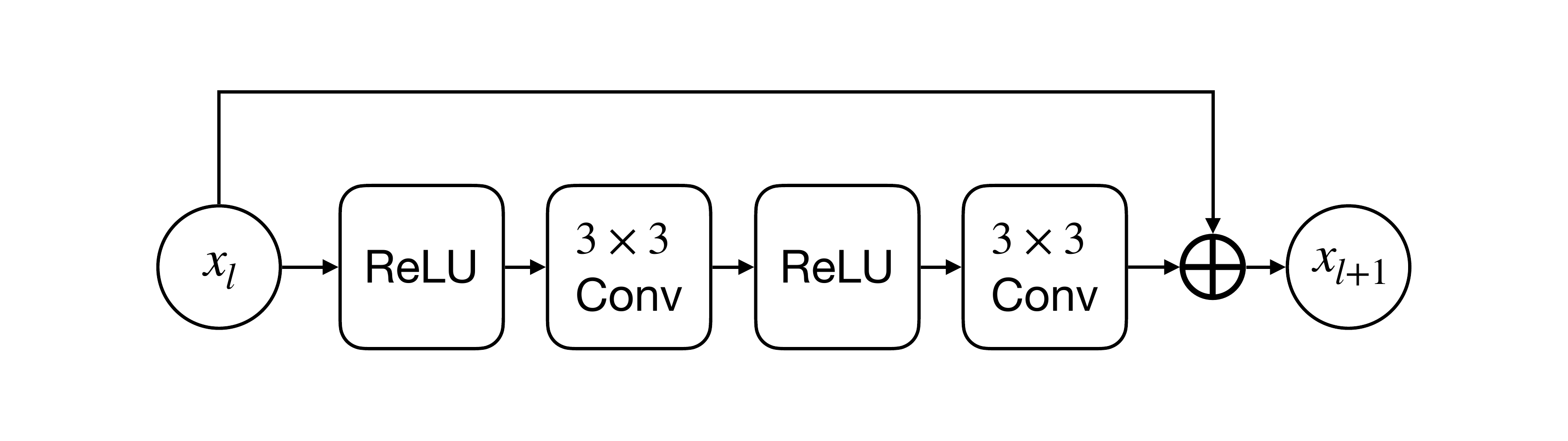}
\end{center}
\caption{The structure of RU in WideResNet. Conv means the standard convolution.}
\label{resblock}
\end{figure}

WideResNet~\cite{zagoruyko2016wide} is a variant of ResNet~\cite{he2016deep} with improved performance; its improved performance is due to the increased number of channels in each layer.
Each layer of WideResNet called \textit{residual unit}~(RU) consists of convolutions and skip connections only~(see Figure~\ref{resblock} for the structure of RU), so that the classification performance of the WideResNet highly depends on the performance of the convolution used in RU. 
Moreover, WideResNet is suitable for applying group convolution techniques since the number of convolutional channels of each layer is very large.
Hence, for numerical experiments that highlight the superiority of the proposed two-level group convolution compared to other group convolution variants, WideResNet is a good choice as a model CNN.
In what follows, we denote WideResNet with hyperparameters $l$ and $w$ by WideResNet-$l$-$w$, where $l$ and $w$ are the number of layers and the channel-widening factor for convolutions in the network, respectively.
We use WideResNet-28-10 and WideResNet-34-2 for the CIFAR and ImageNet datasets, respectively.

\subsubsection{Ablation study}
In order to highlight the effect of the coarse structure in the proposed two-level group convolution, experiments for ablation study were conducted.
We present numerical results of several WideResNet implementations using the standard convolution~(SC), (one-level) group convolution~(GC), and the proposed two-level group convolution~(GC-2L).

\begin{figure}
\begin{center}
\subfigure[WideResNet-28-10]{\includegraphics[width=0.65\linewidth, trim=6cm 0cm 8cm 0cm,clip]{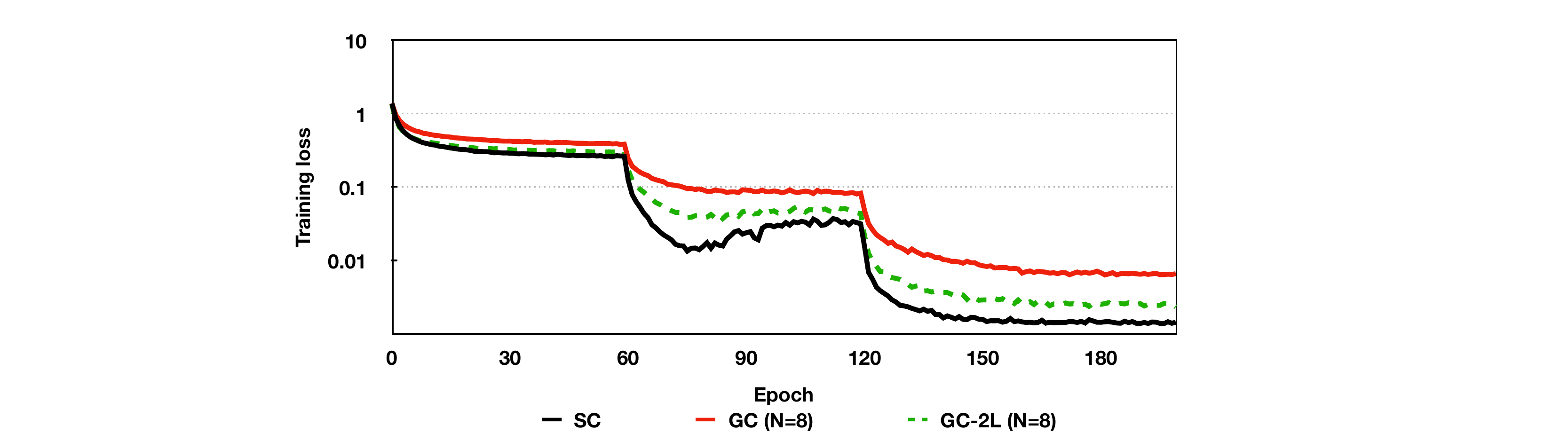}} \\
\subfigure[WideResNet-34-2]{\includegraphics[width=0.65\linewidth, trim=6cm 0cm 8cm 0cm,clip]{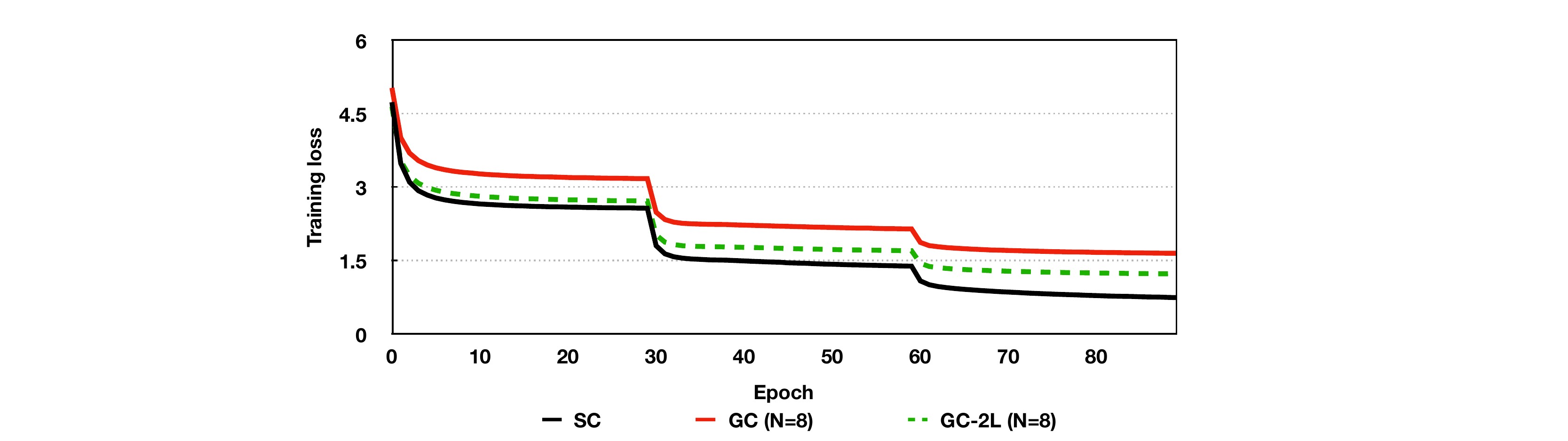}}
\end{center}
\caption{Training loss of WideResNet equipped with the standard convolution~(SC), the group convolution~(GC), and the proposed two-level group convolution~(GC-2L):
\textbf{(a)} WideResNet-28-10 for the CIFAR-10 dataset, \textbf{(b)} WideResNet-34-2 for the ImageNet dataset.}
\label{trloss}
\end{figure}

Figure~\ref{trloss} shows the decay of the training loss of WideResNet equipped with various kinds of convolutions for the CIFAR-10 and ImageNet datasets.
We observe that the training loss of GC-2L converges to a smaller value than GC in both cases.
This phenomenon can be explained mathematically; since GC-2L strictly contains GC in the sense that it reduces to GC when the coarse structure is removed, its minimum loss must be smaller than GC.
Moreover, the loss decay curve of GC-2L lies between those of SC and GC.
It means that the representation capacity of GC-2L is bigger than GC, keeping the parallel structure.

\begin{table}
\begin{center}
 \caption{The error rates, parameters per processor of WideResNet equipped with the standard convolution~(SC), the group convolution~(GC), and the proposed two-level group convolution~(GC-2L)  for the CIFAR and ImageNet datasets,
 for which WideResNet-28-10 and WideResNet-34-2 were used, respectively.}
\label{ablation}
\resizebox{\linewidth}{!}{
\begin{tabular}{lcccc}
 \toprule
 \multicolumn{5}{c}{WideResNet-28-10}\\
 \midrule
 \midrule
 \multirow{2}{*}{Type} & \multirow{2}{*}{$N$}   & \multirow{2}{*}{\begin{tabular}{c}Parameters \\ per processor\end{tabular}}           & \multicolumn{2}{c}{Error rate (\%)}          \\ \cline{4-5}
 &                      &         & CIFAR-10             & CIFAR-100            \\
 \midrule
 SC &1 & 36.48M & 4.19 & 18.95\\
 \midrule
 GC &2 & 18.25M & 4.61 & 20.77\\
 GC-2L&2 & 18.35M & 4.12 & 19.80\\ \hline
 GC &4 & 9.14M & 5.32 & 24.31\\
 GC-2L &4 & 9.25M & 4.34 & 20.90\\ \hline
 GC &8 & 4.58M & 6.54 & 26.89\\
 GC-2L &8 & 4.74M & 4.79 & 22.32\\ \hline
 GC &16 & 2.30M & 8.33 & 30.81\\
 GC-2L &16 & 2.54M & 4.95 & 23.49\\
 \bottomrule
\end{tabular}
\hspace{0.01\textwidth}
\begin{tabular}{lccc}
 \toprule
 \multicolumn{4}{c}{WideResNet-34-2}\\
 \midrule
 \midrule
 \multirow{2}{*}{Type} & \multirow{2}{*}{$N$}   & \multirow{2}{*}{\begin{tabular}{c}Parameters \\ per processor\end{tabular}}           & Error rate (\%)         \\ \cline{4-4}
 &                              &           & ImageNet            \\
 \midrule
 SC &1 & 86.03M & 24.10\\
 \midrule
 GC &2 & 43.55M & 26.75\\
 GC-2L &2 & 43.71M & 25.36\\ \hline
 GC &4 & 22.31M & 30.48\\
 GC-2L &4 & 22.50M & 27.17\\ \hline
 GC &8 & 11.69M & 36.36\\
 GC-2L &8 & 11.95M & 28.43\\ \hline
 GC &16 & 6.38M & 43.18\\
 GC-2L &16 & 6.78M & 29.31\\
 \bottomrule
\end{tabular}
}
\end{center}
\end{table}

Table~\ref{ablation} summarizes numerical results of WideResNet equipped with various kinds of convolutions applied to the CIFAR-10, CIFAR-100 and ImageNet datasets:
the number of parameters per processor and error rate for each convolution according to $N$.
The number of parameters per processor of GC-2L is similar to that of GC, but the error rate is significantly reduced compared to GC.
For example, when $N=16$, WideResNet-34-2 had a $29.31\%$ error rate when equipped with GC-2L, much less than $43.18\%$ of the one-level case.

\subsubsection{Performance of parallel computation}

\begin{table}
\begin{center}
\caption{Numerical results of WideResNet-28-$w$ equipped with SC~($N=1$) and GC-2L~($N=2,4$) for the CIFAR-10 dataset, where $N$ denotes the number of groups.}
\label{scalable}
\begin{tabular}{ccccc}
  \toprule
  $w$ & $N$ & Parameters & Wall-clock time & Error rate (\%)\\
  \midrule
  \multicolumn{1}{l|}{\multirow{3}{*}{40}} & 1   & 583.1M     & 49h 42m 57s & 3.64    \\
  \multicolumn{1}{l|}{}                                  & 2   & 292.0M     & 17h 47m 18s & 3.58    \\
  \multicolumn{1}{l|}{}                                  & 4   & 146.3M     & 8h 39m 47s  &3.73   \\
  \midrule
  \multicolumn{1}{l|}{\multirow{3}{*}{60}} & 1   & 1311.7M     & 89h 49m 09s & 3.51    \\
  \multicolumn{1}{l|}{}                                  & 2   & 656.5M     & 30h 18m 17s & 3.68    \\
  \multicolumn{1}{l|}{}                                  & 4   & 328.7M     & 11h 48m 37s & 3.91    \\
  \midrule
  \multicolumn{1}{l|}{\multirow{3}{*}{80}} & 1   & 2331.8M     & 255h 01m 27s & 3.77    \\
  \multicolumn{1}{l|}{}                                  & 2   & 1166.8M     & 73h 28m 59s & 3.75    \\
  \multicolumn{1}{l|}{}                                  & 4   & 584.0M     & 20h 11m 39s & 4.08    \\
  \bottomrule
\end{tabular}
\end{center}
\end{table}

To check the performance of our two-level group convolution for parallel computation, we conducted experiments while increasing the channel-widening factor $w$ of WideResNet-28-$w$ for CIFAR-10 dataset.
Table~\ref{scalable} shows the error rates and wall-clock times of SC~($N=1$) and GC-2L~($N=2,4$) applied to WideResNet-28-$w$ with $w=40,60$ and $80$.
Note that in the case of $w=60$ and $80$, it was impossible to train WideResNet-28-$w$ equipped with SC with one GPU due to memory problems, so we trained using the naive model parallelism~\cite{lee2014model} with $2$ and $4$ GPUs, respectively.
Also, in the case of GC-2L, the results of $N=2$ and $4$ are computed by utilizing $2$ and $4$ GPUs, respectively.

First, looking at the wall-clock time with $w=40$, the training time of WideResNet-28-$w$ was reduced by about half as $N$ doubled.
In other cases, the training time was reduced by almost a third as $N$ doubled.
In fact, doubling $N$ cuts the number of parameters by half, so the ideal training time is reduced by quarter.
However, it takes more than the ideal time due to the calculation of the coarse structure and communication time.
In addition, the reason why the case of $w=40$ is different from the cases of $w=60, 80$ is that there is a difference in the GPU utilization rate because the number of parameters processed by one GPU is relatively small.
It is natural that as $N$ increases, the number of parameters decreases and the error rate increases accordingly.
But the increase in error rate is very moderate compared to the number of decreased parameters.
This means that GC-2L can effectively reduce the training time with a small increase in error rate when performing parallel computation of convolutions dealing with very large channels.

\subsubsection{Comparison with existing approaches}
\label{wideres_2}
Lastly, we verify that the performance of GC-2L is comparable to existing state-of-the-art variants of group convolution; we compare the performance of GC-2L to two variants of group convolution: ShuffleNet convolution~(Shuffle)~\cite{zhang2018shufflenet} and Fully Learnable Group Convolution~(FLGC)~\cite{wang2019fully}.
Shuffle randomly permutes the channels of the output of a group convolution, resulting in data exchange among groups in the output.
On the other hand, FLGC dynamically determines and updates the input channels and filters for each group according to the overall loss of the network.
Both Shuffle and FLGC lose the parallel structure of group convolution due to large data exchange among groups so that it may not be efficient to implement them in a distributed memory system because of communication bottlenecks.
Also, the authors of Shuffle and FLGC did not apply the proposed convolution to all convolutional layers of the network, but only a specific part.
However, for fair comparison, we applied the proposed convolutions to all convolutional layers and tested them.
In the following, we present numerical results executed on a single GPU in order to highlight the performance of our proposed method compared to Shuffle and FLGC.

\begin{table}
\begin{center}
\caption{The error rates and total number of parameters of WideResNet equipped with the group convolution~(GC), two-level group convolution(GC-2L), Shuffle, Fully Learnable Group Convolution~(FLGC) for the CIFAR and ImageNet datasets, where $N$ denotes the number of groups.}
\label{error1}
\resizebox{\linewidth}{!}{
\begin{tabular}{lcccc}
 \toprule
 \multicolumn{5}{c}{WideResNet-28-10}\\
 \midrule
 \midrule
 \multirow{2}{*}{Type} & \multirow{2}{*}{$N$}   & \multirow{2}{*}{Parameters}           & \multicolumn{2}{c}{Error rate (\%)}          \\ \cline{4-5}
 &                      &         & CIFAR-10             & CIFAR-100            \\
 \midrule
 GC &2 & 18.25M & 4.61 & 20.77\\
 Shuffle &2 & 18.25M & 4.00 & 19.51\\
 FLGC &2 & 18.25M & 4.62 & 21.68\\
 GC-2L&2 & 18.35M & 4.12 & 19.80\\
 \midrule
 GC &4 & 9.14M & 5.32 & 24.31\\
 Shuffle &4 & 9.14M & 4.21 & 20.38\\
 FLGC &4 & 9.14M & 6.16 & 26.33\\
 GC-2L &4 & 9.25M & 4.34 & 20.90\\
 \midrule
 GC &8 & 4.58M & 6.54 & 26.89\\
 Shuffle &8 & 4.58M& 4.73 & 21.91\\
 FLGC &8 & 4.58M & 9.81 & 30.78\\
 GC-2L &8 & 4.74M & 4.79 & 22.32\\
 \midrule
 GC &16 & 2.30M & 8.33 & 30.81\\
 Shuffle &16 & 2.30M & 5.28 & 24.20\\
 FLGC &16 & 2.30M  & 11.26 & 37.49\\
 GC-2L &16 & 2.54M & 4.95 & 23.49\\
 \bottomrule
\end{tabular}
\hspace{0.05\textwidth}
\begin{tabular}{lccc}
 \toprule
 \multicolumn{4}{c}{WideResNet-34-2}\\
 \midrule
 \midrule
 \multirow{2}{*}{Type} & \multirow{2}{*}{$N$}   & \multirow{2}{*}{Parameters}           & Error rate (\%)         \\ \cline{4-4}
 &                              &           & ImageNet            \\
 \midrule
 GC &2 & 43.55M & 26.75\\
 Shuffle &2 & 43.55M & 25.12\\
 FLGC &2 & 43.55M & 30.43\\
 GC-2L &2 & 43.71M & 25.36\\
 \midrule
 GC &4 & 22.31M & 30.48\\
 Shuffle &4 & 22.31M & 26.84\\
 FLGC &4 & 22.31M & 38.30\\
 GC-2L &4 & 22.50M & 27.17\\
 \midrule
 GC &8 & 11.69M & 36.36\\
 Shuffle &8 & 11.69M & 29.03\\
 FLGC &8 & 11.69M & 45.54\\
 GC-2L &8 & 11.95M & 28.43\\
 \midrule
 GC &16 & 6.38M & 43.18\\
 Shuffle &16 & 6.38M & 32.27\\
 FLGC &16 & 6.38M & 62.02\\
 GC-2L &16 & 6.78M & 29.31\\
 \bottomrule
\end{tabular}
}
\end{center}
\end{table}

Table~\ref{error1} shows numerical results of all convolutions mentioned above applied to WideResNet-28-10 with the CIFAR-10, CIFAR-100 datasets and WideResNet-34-2 with the ImageNet dataset.
The number of total parameters of GC-2L is slightly larger than those of Shuffle and FLGC; such a difference is due to the coarse structure of GC-2L.
The classification performance was recorded in terms of the error rate.
The error rate of GC-2L is less than $5\%$ for CIFAR-10, $24\%$ for CIFAR-100 and $30\%$ for ImageNet.
The error rates of GC-2L are generally similar to or even lower than other variants.

\begin{figure}
\begin{center}
\subfigure[WideResNet-28-10]{\includegraphics[width=0.65\linewidth, trim=6cm 0cm 8cm 0cm,clip]{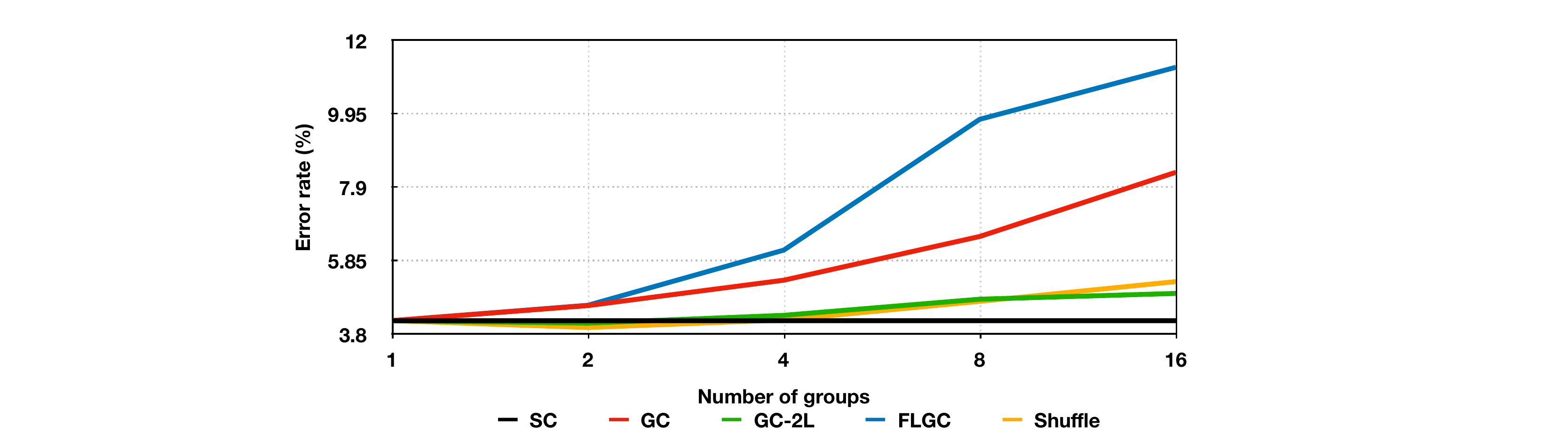}} \\
\subfigure[WideResNet-34-2]{\includegraphics[width=0.65\linewidth, trim=6cm 0cm 8cm 0cm,clip]{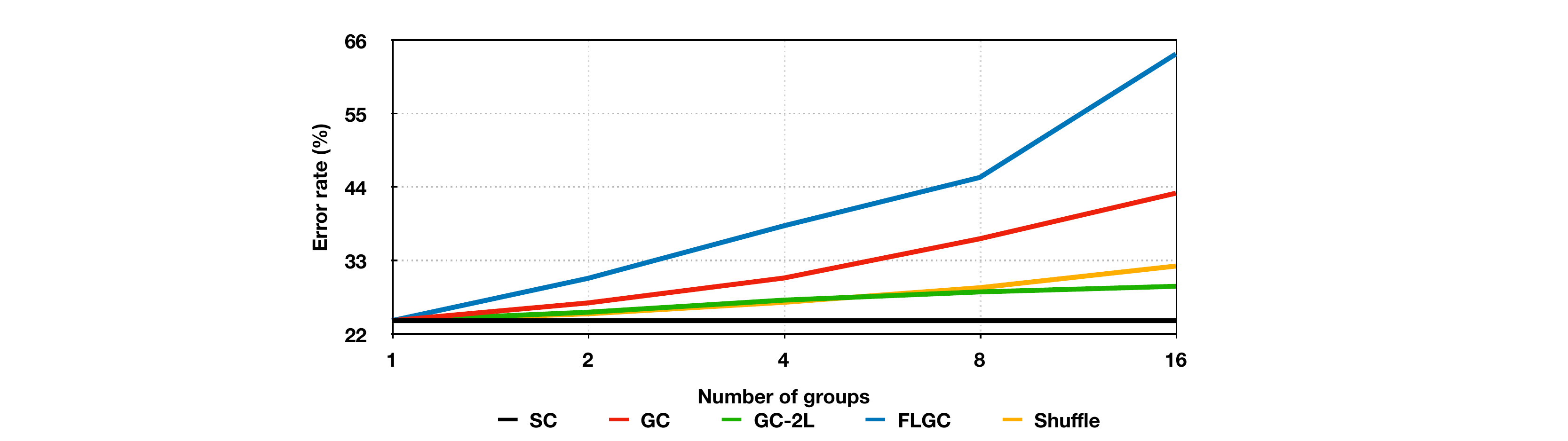}}
\end{center}
\caption{The error rate graph of WideResNet equipped with SC, GC, GC-2L, FLGC, and Shuffle:
\textbf{(a)} WideResNet-28-10 for the CIFAR-10 dataset, \textbf{(b)} WideResNet-34-2 for the ImageNet dataset.}
\label{errorgraph_w}
\end{figure}

Moreover, Figure~\ref{errorgraph_w} shows that GC-2L has the lowest slope of error rate increase even as the number of groups increases.
Also, in both cases, Shuffle and GC-2L showed similar performance until $N \le 8$, but when $N>8$, GC-2L showed better performance.
What this means is that the channel shuffling technique showed good performance in resolving the lack of inter-group communication when $N$ is small. However, when $N$ is large, a suitable coarse structure is required rather than simple shuffling.
On the other hand, FLGC showed the poor performance compared to others in our experiments; FLGC was originally proposed with $1 \times 1$ convolution and it seems that FLGC do not show the good performance with convolutions of large kernel size.

\subsection{MobileNetV2}
\label{Sec:Mobile}

\begin{figure}
\begin{center}
\includegraphics[width=0.8\linewidth, trim=0cm 0cm 0cm 0cm,clip]{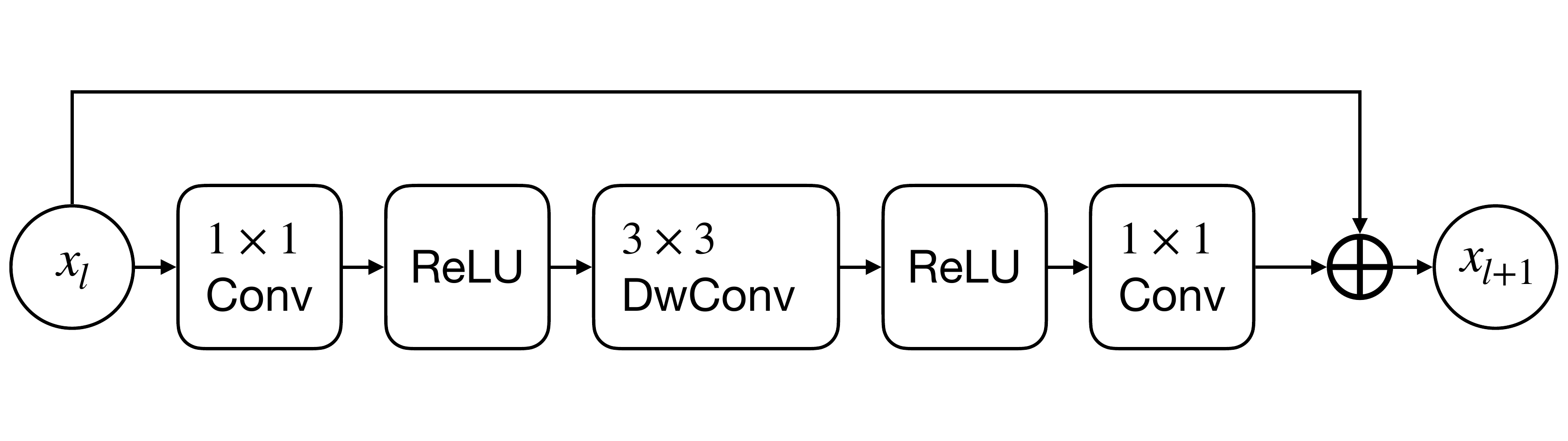}
\end{center}
\caption{The structure of an inverted residual block in MobileNetV2. Here, Conv and DwConv mean standard convolution and depthwise convolution, respectively.}
\label{inverted}
\end{figure}

MobileNetV2~\cite{sandler2018mobilenetv2} is a CNN architecture designed for use in smartphones; its number of parameters is relatively small so that it is suitable to lightweight computer architectures such as mobile devices.
The core structure of MobileNetV2 is an \textit{inverted residual block}. 
It uses $1 \times 1$ standard convolutions to increase or decrease the number of channels.
Figure~\ref{inverted} shows the structure of the inverted residual block in MobileNetV2.
The main computational cost of the inverted residual block is due to two $1 \times 1$ standard convolutions.
Hence, similar to the case of $3 \times 3$ convolutions in WideResNet, group convolution and its variants can be utilized to reduce the computational cost of $1 \times 1$ standard convolutions in the inverted residual block.
In particular, we note that Shuffle and FLGC were originally proposed for application to $1 \times 1$ convolutions~\cite{wang2019fully,zhang2018shufflenet}.
Based on this inverted residual block architecture, we replace the $1 \times 1$ convolution layers with GC, GC-2L, Shuffle, and FLGC.

\begin{table}
\begin{center}
\caption{The error rates and total number of parameters of MobileNetV2 equipped with the standard $1 \times 1$ convolution~(SC), group convolution~(GC), Shuffle, Fully Learnable Group Convolution~(FLGC), and the proposed two-level group convolution~(GC-2L) for the ImageNet dataset, where $N$ denotes the number of groups.}
\label{error3}
\begin{tabular}{lccc}
  \toprule
  \multicolumn{4}{c}{MobileNetV2}\\
  \midrule
  \midrule
  \multirow{2}{*}{Type} & \multirow{2}{*}{$N$}   & \multirow{2}{*}{Parameters}           & Error rate (\%)         \\ \cline{4-4}
  &                              &           & ImageNet            \\
  \midrule
  SC &1 & 3.50M & 30.61\\
  \midrule
  GC &2 & 2.44M & 36.96\\
  Shuffle &2 & 2.44M & 36.67\\
  FLGC &2 & 2.44M & 37.98\\
  GC-2L &2 & 2.47M& 33.56\\
  \midrule
  GC &4 & 1.91M & 42.63\\
  Shuffle &4 & 1.91M & 38.14\\
  FLGC &4 & 1.91M & 44.37\\
  GC-2L &4 & 1.95M& 36.88\\
  \midrule
  GC &8 & 1.65M & 50.36\\
  Shuffle &8 & 1.65M & 42.75\\
  FLGC &8 & 1.65M & 49.28\\
  GC-2L &8 & 1.72M & 39.16\\
  \bottomrule
\end{tabular}
\end{center}
\end{table}

\begin{figure}
\begin{center}
\includegraphics[width=0.65\linewidth, trim=6cm 0cm 8cm 0cm,clip]{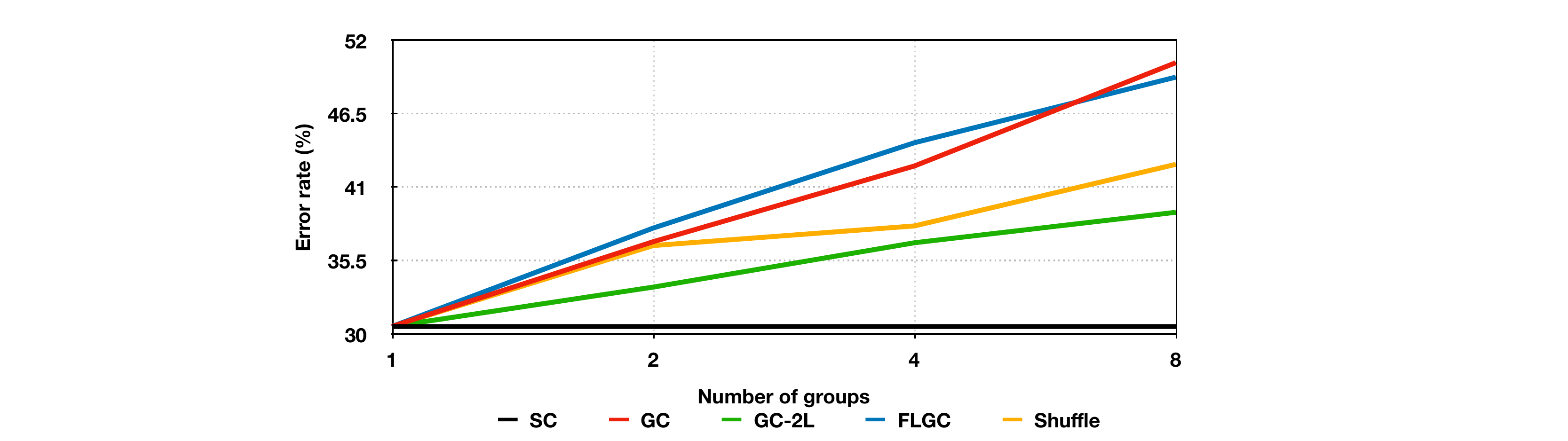}
\end{center}
\caption{The error rate graph of MobileNetV2 equipped with SC, GC, GC-2L, FLGC, and Shuffle for the ImageNet dataset.}
\label{errorgraph_m}
\end{figure}

Numerical results in Table~\ref{error3} ensure that the proposed two-level group convolution effectively reduces the computational cost of $1 \times 1$ convolution without severe performance degradation.
The number of parameters of GC-2L are slightly greater than those of other group convolution variants, but still considerably less than that of standard convolution.
In view of the classification performance, we can observe that the error rate of GC-2L is the lowest among all group convolution variants.
In particular, the error rate of GC-2L is less than $40\%$ even if the number of groups increases to $8$; this highlights the robustness of GC-2L for multi-GPU computing compared to other state-of-the-art group convolution variants.
Figure~\ref{errorgraph_m} shows the increase rate of the error of each convolution as the number of groups increases.
One can readily observe that the graph corresponding to GC-2L has the smallest slope among all the graphs of the group convolution variants.
Hence, we can expect that the greater the numeber of groups, the greater the difference between the error rate of GC-2L and other error rates.
We conclude that adding an appropriate coarse structure to the group convolution is a more effective strategy than shuffling channels in the sense of  both performance issue and distributed memory computing.

\section{Conclusion}
\label{Sec:Conc}
In this paper, we proposed a novel approach for group convolution that is suitable for distributed memory computing and robust with respect to the large number of groups.
Motivated by two-level methods in numerical analysis, a coarse structure that promotes intergroup communication was added to the group convolution.
All parameters of the coarse structure can be distributed among multiple GPUs so that the proposed two-level group convolution is suitable for parallel computation.
We presented how the additional coarse structure helped the improvement of the performance of the group convolution through an ablation study.
We also observed that the two-level group convolution performed similarly or better than conventional variants of group convolution.
We expect that the proposed two-level group convolution can be efficiently utilized for training CNNs with a large number of channels, such as WideResNet, through multiple GPUs.
Additionally, the two-level group convolution is expected to be effective for parameter pruning of small networks for smartphones such as MobileNetV2.

\section*{Declaration of Competing Interest}
The authors declare that they have no known competing financial interests or personal relationships that could have appeared to influence the work reported in this paper.

\section*{Acknowledgments}
This work was supported in part by the National Research Foundation~(NRF) of Korea grant funded by the Korea government (No.~2020R1A2C1A01004276), and in part by Basic Science Research Program through NRF funded by the Ministry of Education (No.~2019R1A6A1A10073887).
\bibliography{mybibfile}

\end{document}